\documentclass[letterpaper]{article} 
\usepackage{aaai25}  
\usepackage{times}  
\usepackage{helvet}  
\usepackage{courier}  
\usepackage[hyphens]{url}  
\usepackage{graphicx} 
\usepackage{natbib}  
\usepackage{caption} 
\usepackage{algorithm}

\usepackage{newfloat}
\usepackage{listings}
\DeclareCaptionStyle{ruled}{labelfont=normalfont,labelsep=colon,strut=off} 
\floatstyle{ruled}
\newfloat{listing}{tb}{lst}{}
\floatname{listing}{Listing}
\usepackage{amsfonts} 
\usepackage{algpseudocode}
\usepackage{amsmath} 
\usepackage{booktabs} 
\usepackage{subcaption}

\title{ParseCaps: An Interpretable Parsing Capsule Network \\ for Medical Image Diagnosis}
\author{
    Xinyu Geng\textsuperscript{\rm 1},
    Jiaming Wang\textsuperscript{\rm 1},
    Jun Xu\textsuperscript{\rm 1}}
\affiliations{
    \textsuperscript{\rm 1}School of Mechanical Engineering and Automation

    Harbin Institute of Technology, Shenzhen \\
    22s153095@stu.hit.edu.cn, 21s153144@stu.hit.edu.cn, xujunqgy@hit.edu.cn
}

\usepackage{bibentry}

\begin{document}


\maketitle

\begin{abstract}

Deep learning has excelled in medical image classification, 
but its clinical application is limited by poor interpretability. 
Capsule networks, known for encoding hierarchical relationships and spatial features, 
show potential in addressing this issue. 
Nevertheless, traditional capsule networks often underperform due to their shallow structures,   
and deeper variants lack hierarchical architectures, thereby compromising interpretability. 
This paper introduces a novel capsule network, ParseCaps, 
which utilizes the sparse axial attention routing and parse convolutional capsule layer to form a parse-tree-like structure, 
enhancing both depth and interpretability. 
Firstly, sparse axial attention routing optimizes connections between child and parent capsules, 
as well as emphasizes the weight distribution across instantiation parameters of parent capsules.  
Secondly, the parse convolutional capsule layer generates capsule predictions aligning with the parse tree. 
Finally, based on the loss design that is effective whether concept ground truth exists or not, 
ParseCaps advances interpretability by associating each dimension of the global capsule with a comprehensible concept, 
thereby facilitating clinician trust and understanding of the model's classification results.
Experimental results on CE-MRI, PH$^2$, and Derm7pt datasets show 
that ParseCaps not only outperforms other capsule network variants in classification accuracy, redundancy reduction and robustness,  
but also provides interpretable explanations, regardless of the availability of concept labels.

\end{abstract}

%

\section{Introduction}

Deep learning methods of medical image classification 
provides consistent, rapid predictions
that often exceed human capabilities in detecting subtle abnormalities.
However, obtaining extensive, high-quality datasets is challenging  
and poor interpretability limits their clinical application.
Capsule networks (CapsNets) have shown potential to enhance interpretability 
by maintaining hierarchical relationships and spatial orientations within images \cite{sabour2017dynamic,hinton2018matrix}. 
CapsNets not only excel in generalizing from small datasets, 
but also improve classification accuracy by capturing relationships between disease markers 
and normal anatomical structures \cite{akinyelu2022brain,ribeiro2022learning,patrick2022capsule}. 
CapsNets utilize vectors called capsules to replace single neuron.
Each capsule vector's length represents the presence probability of specific entity in the input image,
and its direction encodes the captured features \cite{sabour2017dynamic}. 
Each dimension of the capsule vector, termed an instantiation parameter, 
represents the direction of the capsule, 
thus conferring inherent physical meanings.
These parameters hold potential for concept interpretability, 
as each corresponds to a human-understandable and meaningful concept.
Detailed introduction of CapsNets is in supplementary material \ref{5.3}.

Existing CapsNets face challenges to assign clear meaning to instantiation parameters; 
however, integrating a parse-tree-like structure 
could map part-to-whole relationships similarly to human cognitive processes \cite{sabour2017dynamic},
thereby enhancing concept interpretability.
In this structure, each node is a capsule, and through routing, 
active capsules select parent capsules from the upper layer. 
Because \cite{sabour2017dynamic} does not adhere to a strict tree structure where each child node connects to only one parent node, 
it is referred to as parse-tree-like.
The top layer features a single ``global capsule" that provides a comprehensive view of the entire image, 
encapsulating entities and their hierarchical relationships. 
This allows each instantiation parameters to align with an interpretable and conceptually meaningful image entity. 
\textbf{The parse tree provides a carrier for interpretability through the global capsule. 
Coupled with loss constraints, 
they create a capsule network with concept interpretability.}

However, the current implementation of parse tree faces challenges. 
First, it lacks a suitable routing algorithm that effectively supports the parse-tree-like structure.
Dynamic routing process tends to create a fully connected structure between sub-capsules and parent capsules   
\cite{peer2018training,jeong2019ladder}, 
which undermines the desired selective connection essential for clear hierarchical relationships.
Additionally, although existing attention routing method sparsifies the coupling coefficients 
between capsules \cite{geng2024orthcaps}, 
it does not see the significance of instantiation parameters within capsules, 
which adversely affects the global capsule's instantiation parameters.
Second, certain CapsNet layers contradict the parse tree 
by increasing the number of capsules without correspondingly enhancing their dimension \cite{rajasegaran2019deepcaps,choi2019attention,geng2024orthcaps}., 
which is detailed in Sec. \ref{3.2.1}.
They fail to concentrate features of entire image into a global capsule; 
instead, as the number of capsules increases, features become more dispersed,
leading to redundancy in feature representation and a diluted hierarchical structure.
Furthermore, existing CapsNets lack loss functions to promote concept interpretability.

\begin{figure}
    \vspace*{-10pt}
    \centering
    \includegraphics[width=0.75\columnwidth]{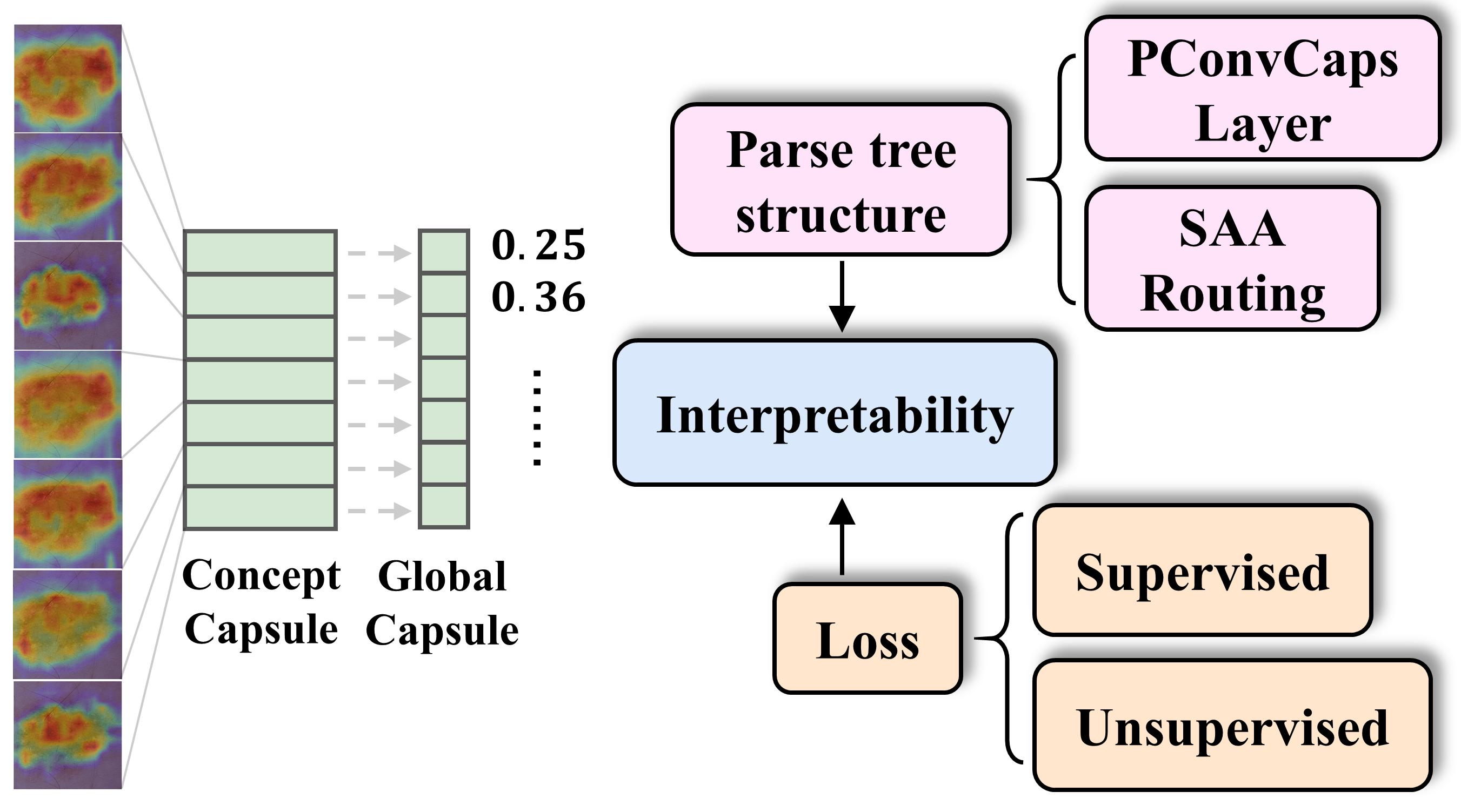} 
    \caption{The visualization of motivation. Using a parse-tree-like structure as a carrier, combined with the constraints of loss, an interpretable capsule network with instantiation parameters has been achieved. Each dimensional instantiation parameter of the global capsule corresponds to a human-understandable concept.}
    \label{Intro}
    \vspace*{-12pt}
\end{figure}

This paper presents ParseCaps, a novel capsule network featuring three enhancements corresponding to above challenges. 
First, following \cite{geng2024orthcaps}, a sparse axial attention (SAA) routing is proposed,  
which not only sparsifies coupling coefficients
but also weighs instantiation parameters within each capsule.  
The sparsity inherent in SAA routing 
restricts routing process, excluding sub-capsules with weaker connections to parent capsules, 
thus benefiting a hierarchical parse-tree-like structure. 
Second, 
we introduce a parse convolutional capsule (PConvCaps) layer. 
This layer generates capsules whose prediction strictly aligns with the parse tree, 
reducing the number of capsules while increasing their dimensionality as layer depth increases, 
thus forming an interpretable global capsule.
Third, we design loss functions to enable the model to provide conceptual explanations 
that aligns each dimension of the capsule vector with a human-understandable concept,
regardless of whether concept ground truth are available or not.
The motivation of this paper is shown in Fig. \ref{Intro}.

\paragraph{Contributions}
\textbf{1)} ParseCaps forms a parse-tree-like structure by
creating PConvCaps layer and SAA routing to optimize the network's hierarchical structure, 
especially focusing on instantiation parameter weights in SAA routing.
\textbf{2)} It enhances interpretability in CapsNets with a parse-tree-like structure and loss functions, 
aligning instantiation parameters of global capsule with human-understandable concepts.
\textbf{3)} It outperforms existing CapsNets in medical image classification and provides interpretable explanations
regardless of concept label availability.

\section{Related work}

\paragraph{Capsule networks and parse-tree-like structure}
\cite{hinton1999learning} first proposed the parse-tree-like structure for image processing , 
where lower-level nodes represent parts of an entity (i.e., eyes, nose, etc.),  
and top-level nodes depict the entire entity (i.e., the entire face).
Most existing CapsNets with parse trees are constructed within shallow networks.
\cite{sabour2017dynamic} introduced CapsNet with a basic two-layer parse-tree-like structure. 
\cite{peer2018training} proposed a dynamic parse tree for a four-layer CapsNet, 
and \cite{bui2021treecaps} utilized shared weights routing for constructing syntax trees in code comprehension tasks, 
leading to a two-layer capsule network . 
\cite{yu2022hp} explored unsupervised facial parsing with a two-layer capsule encoder. 
However, deeper capsule networks depend on convolutional layers disrupt the parse-tree-like structure \cite{rajasegaran2019deepcaps,sun2020deep,everett2023vanishing}. 
Although \cite{lalonde2020encoding} investigate the interpretability of capsule networks in the medical field, 
it did not address the conceptual significance of instantiation parameters or the parse tree structure.
In conclusion, 
the potential for interpretability and parsing structure in deep capsule networks remains underexplored.

\paragraph{Explainable medical image classification}
Despite the excellent performance of existing deep learning algorithms, their clinical deployment remains limited
primarily due to the non-transparency of their decision-making processes \cite{patricio2023explainable}, 
often described as ``black-box" models \cite{lipton2017doctor}. 
This encourages the development of interpretable deep learning algorithms 
that combine explainable models with high medical diagnostic accuracy \cite{gunning2019darpa}.
Early methods of interpretability involved perturbing input images to see how model outputs changed \cite{zeiler2014visualizing,ribeiro2016should,lundberg2017unified}, 
or analyzing model activations to determine essential lesions for predictions \cite{zhou2016learning,selvaraju2017grad}. 
However, these post-hoc explanation methods were criticized for potentially missing the complex dynamic process within models \cite{adebayo2018sanity,rudin2019stop}. 
Consequently, there is a growing trend towards designing models
that inherently explain their decision-making processes \cite{kim2021xprotonet,wickramanayake2021comprehensible,gallee2023interpretable}. 
\cite{alvarez2018towards,sarkar2022framework} constructed models with built-in, pre-hoc interpretability, 
known as conceptual interpretability, 
which clarified classifier predictions through a set of human-understandable concepts.

\section{Methodology} \label{Methodology}

The interpretability of ParseCaps stems from the parse-tree-like structure and loss functions. 
Sec. \ref{3.1} outlines the overall model structure, Sec. \ref{3.2} details the construction of the parse tree, 
and Sec. \ref{3.3} describes loss function design. All symbols are summarized in supplementary material \ref{5.1}.

\subsection{Overall architecture} \label{3.1}

\begin{figure*}
  \vspace*{-15pt}
    \centering
    \includegraphics[width=0.8\textwidth]{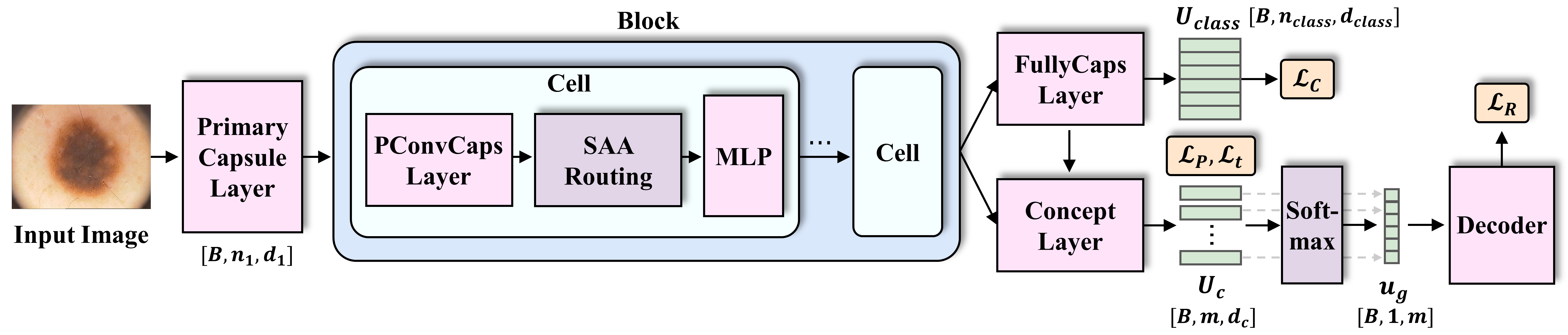} 
    \vspace*{-10pt}
    \caption{The architecture of ParseCaps.}
    \label{A}
    \vspace*{-12pt}
\end{figure*}

The ParseCaps architecture is shown in Fig. \ref{A}.
The input image \( x \) is fed into 
the initial convolutional block consisting of four convolutional layers that
extract features $\Phi^0 \in \mathbb{R}^{(B, f_0, W_0, H_0 )}$,
where \( B \), \( f_0 \), \( W_0 \), and \( H_0 \) denote the batch size, number of feature maps, width and height, respectively.
This block maps the image features to a higher dimentional space, 
facilitating capsule creation \cite{mazzia2021efficient}.
Then, the primary capsule layer converts $\Phi^0$ into capsules \( U_1 \in \mathbb{R}^{(B, n_1, d_1)} \),
where \( n_1 \) and \( d_1 \) are the number and dimension of primary capsules, respectively.

A capsule block consists of several capsule cells, 
each containing a PConvCaps layer, SAA routing, and MLP. 
The PConvCaps layer generates predictions \( \hat{U}_{l+1} \) 
for higher-level capsules from lower-level \( U_l \). 
SAA routing computes coupling coefficients $c_{ij}$ based on the alignment between \( U_l \) and \( \hat{U}_{l+1} \). 
The MLP deepens the network without changing capsule counts or dimensions. 
Capsule blocks can be stacked to extend the network depth effectively. 
ParseCaps includes three blocks with one, two, and five cells, respectively.

The final block connects to FullyCaps layer and concept layer, 
the FullyCaps layer converges all capsules into class capsules $U_{class} \in \mathbb{R}^{(B, n_{class}, d_{class})}$ to make predictions.
In concept layer, inputs are mapped into \( m \) concept capsules $U_{c} \in \mathbb{R}^{(B, m, d_c)}$ by two linear layers.
Each $u_{c,i}$ is converted into a 1D instantiation parameter of the global capsule $u_g \in \mathbb{R}^{(B, 1, m)}$,
representing $m$ concepts $p_1, p_2, \dots, p_m$. 
The softmax function renders each instantiation parameter to $[0,1]$ as the activation of that concept.
$u_g$ feeds into a decoder with four deconvolution layers to reconstruct \( x \).

\subsection{Parse-tree-like structure} \label{3.2}

\cite{peer2018training} identified two essential criteria for a parse tree structure :  
First, there should be fewer parent capsules than sub-capsules, 
with parent capsules having greater dimensions to encapsulate more features.
Second, each sub-capsule should connect to only one parent capsule. 
PConvCaps layer meets the first criterion by
buliding a model structure that widens at the base and narrows at the top.
For the second criterion, 
it causes significant feature loss and overfitting to maintain one-to-one connections between parent capsules 
and sub-capsules.
SAA routing is used to reduce unnecessary connections.

\subsubsection{Parse convolutional capsule layer} \label{3.2.1}

Recent studies on CapsNets commonly employ convolutional capsule (ConvCaps) layer \cite{rajasegaran2019deepcaps,geng2024orthcaps,choi2019attention}.
The input of the ConvCaps layer $l$ is $U_l \in \mathbb R^{(B,w_l,w_l,d_l,n_l)}$, 
transforming into the output $\hat U_{l+1} \in \mathbb R^{(B,w_{l+1},w_{l+1},d_{l+1},n_{l+1})}$, 
where $B, w_l, d_l$ and $n_l$ are the batchsize, width of the input feature map,
the dimension and the number of capsules, respectively.
When performing convolution,
$U_l$ is reshaped into a 4D tensor $U_l \in \mathbb R^{(B,w_l,w_l,d_l \times n_l)}$, 
where $d_l \times n_l$ serves as the channel dimension of the tensor.
As the convolutional layers deepen, $d_l \times n_l$ increases while $w_l$ decreases. 
Typically, $d_l$ is often set as a constant, thus the number of capsules $n_l$ increases.

This phenomenon creates challenges in CapsNets.
First, the parse-tree-like structure prefers that 
lower-layer capsules should be short (i.e. small vector dimension $d$) 
and numerous (i.e. large vector count $n$) to represent partial features, 
higher-layer capsules are fewer but longer, encompassing entire image entities.
However, ConvCaps layers fail to adhere to this structure. 
Furthermore, increasing capsule counts without expanding feature space
results in an overabundance of feature carriers.
Capsules represent overlapping areas of feature distribution, leading to redundancy.
Lastly, the 5D capsule \( (B,w_l,w_l,d_l,n_l) \)
combines feature maps and capsules, 
which complicates the meaning of instantiation parameters and reduces interpretability.

\begin{figure}
    \centering
    \includegraphics[width=0.7\columnwidth]{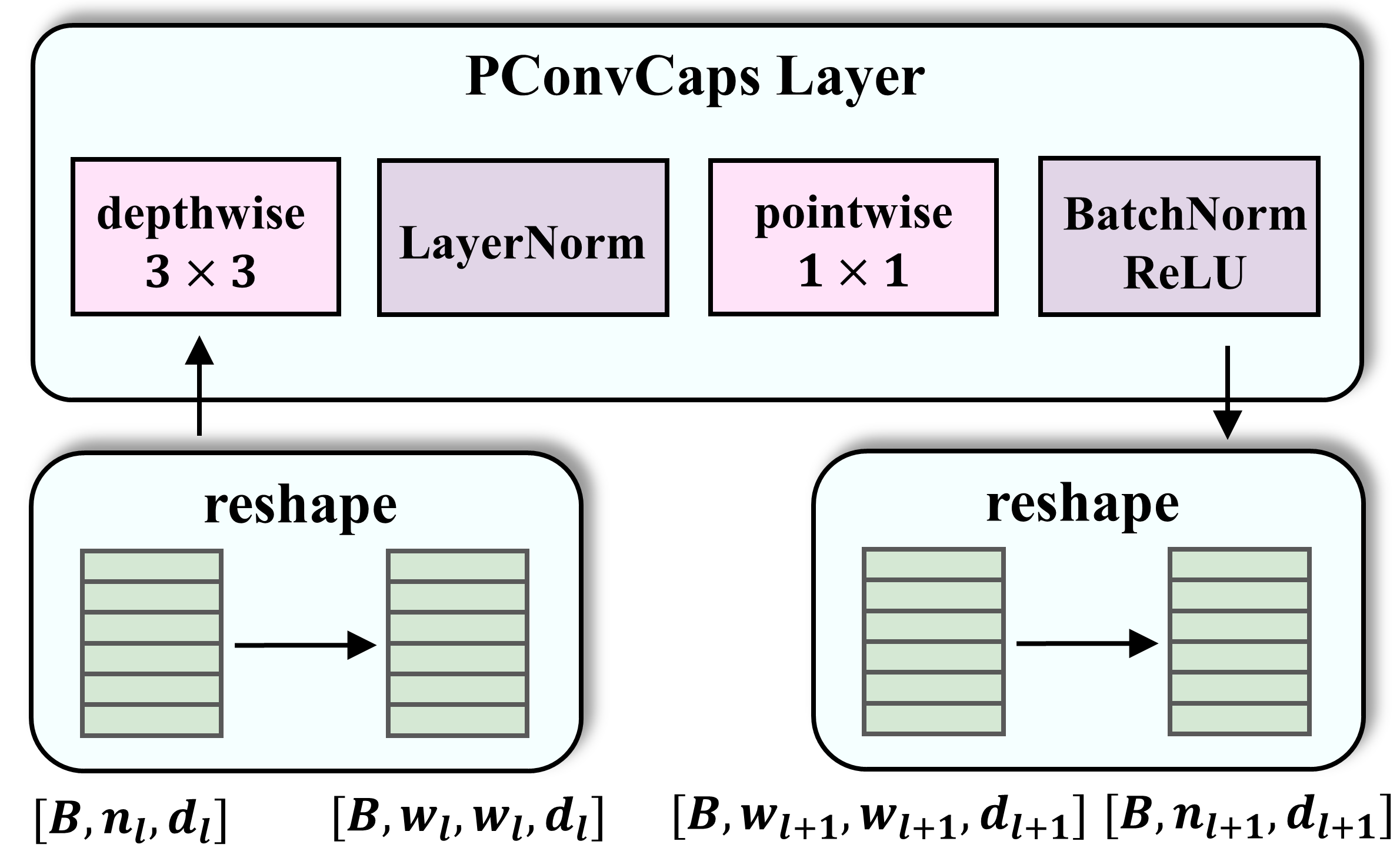} 
    \vspace*{-3pt}
    \caption{The process of PConvCaps Layer.}
    \vspace*{-15pt}
    \label{ConvCaps}
\end{figure}

We introduce the PConvCaps Layer detailed in Fig. \ref{ConvCaps}. 
This layer ensures that all extracted features are represented by capsules, 
with the input capsule defined as $U_l \in \mathbb{R}^{(B, n_l, d_l)}$. 
To adapt to convolution, \( n_l \) is split into \( \sqrt{n_l} \times \sqrt{n_l} \), 
and \( d_l \) corresponds to the number of channels in the convolutional layer.
We define the width of the spatial grid be \( w_l = \sqrt{n_l} \),
and reshape $U_l$ into \( (B, w_l, w_l, d_l) \).

Consequently, when the stride is larger than 1, 
the number of capsules \( n_l \) naturally decreases 
and the dimension \( d_l \) increases as the layers deepen, aligning with the parse tree. 
PConvCaps layer uses depthwise convolution \cite{chollet2017xception} 
with a $3 \times 3$ kernel and stride of 2,
followed by layer normalization \cite{ba2016layer} 
and pointwise convolution with a $1 \times 1$ kernel and stride of 1.
This is a lightweight convolution operation that reduces parameters.

\subsubsection{Sparse axial attention routing} \label{3.2.2}

We develop the sparse axial attention routing (SAA routing) 
which facilitates the parse-tree-like structure 
and focuses on the weight distribution of the instantiation parameters.
SAA routing consists of two parts: sparse attention and axial attention.

\paragraph{Sparse attention}

\begin{figure}
  \vspace*{-10pt}
  \centering
  \includegraphics[scale=0.08]{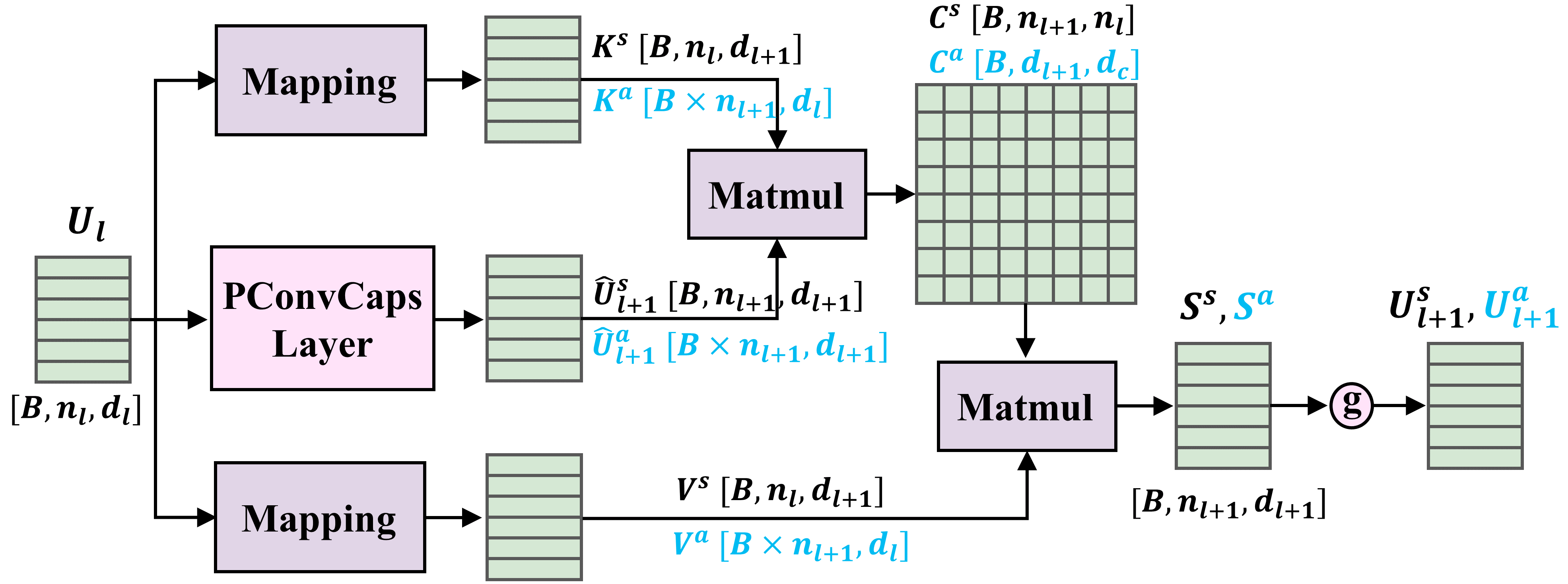} 
  \vspace*{-10pt}
  \caption{The process of SAA routing. The tensor shapes for sparse attention and axial attention are represented in black and blue, respectively.
  $g$ is the activation function Squash.}
  \label{SAA}
  \vspace*{-12pt}
\end{figure}

The essence of sparse attention routing is capturing the coupling coefficients $C^{s}$
between lower-level capsules and higher-level capsules via an attention map,
as shown in Fig. \ref{SAA}, where $s$ denotes the sparse attention.
For the \(l\)-th layer capsules \( U_{l} \in \mathbb{R}^{(B , n_l , d_l)} \), 
the PConvCaps layer generates predictions of \((l+1)\)-th layer capsules \( \hat U_{l+1}^{s} \in \mathbb{R}^{(B, n_{l+1}, d_{l+1})} \),
consisting of $n_{l+1}$ capsules, each with $d_{l+1}$ dimensions.
$\hat U_{l+1}^{s}$, \( K^{s} \in \mathbb{R}^{(B, n_{l}, d_{l+1})} \) and \( V^{s} \in \mathbb{R}^{(B, n_{l}, d_{l+1})} \) 
serve as the query, key and value in the attention mechanism, respectively.
For matrix multiplication, \( K^{s} \) and \( V^{s}\) are mapped to $(B,n_l,d_{l+1})$    
through pointwise convolution with a kernel size of \( 1 \times 1 \) and stride 1.
The computation of attention scores 
corresponds to calculate coupling coefficients $c_{ij}$ in dynamic routing.
The coupling coefficient matrix \( C^{s} \in \mathbb{R}^{(B, n_{l+1}, n_{l})} \) is calculated in Eq. \ref{eq3}:
\begin{equation}
     C^{s} = \alpha\text{-Entmax}( \frac {\hat U_{l+1}^{s} {(K^{s})}^{\rm{T}}}{\sqrt{d_{l+1}}})
    \label{eq3}
\end{equation}
The dot product is scaled by dividing \( \sqrt{d_{l+1}} \) to prevent excessively high attention scores, 
thereby stabilizing the gradients. 
\( \alpha\text{-Entmax} \) is the sparse softmax \cite{correia2019adaptively} and is defined as Eq. \ref{a_entmax}:
\begin{equation}
  \alpha\text{-Entmax}(x)_i = \max \left( \frac{x_i - \tau}{\alpha}, 0 \right)^{\frac{1}{\alpha - 1}}
  \label{a_entmax}
\end{equation}
$\tau$ is a self-adaption threshold and $\alpha$  
controls the sparsity of the attention map.
$\alpha$-Entmax adaptively sets smaller coupling coefficient \( c_{ij} \) to zero, 
encouraging sub-capsules to connect to most relevant parent capsules. 
This selective connectivity is crucial for the parse-tree-like structure.
After assigning \( c_{ij} \) to \( \hat{U}_{l+1}^{s}  \), 
the votes \( S^{s} \) are generated in Eq. \ref{eq4}:
\begin{equation}
    S^{s} = C^{s} {(V^{s})}^{\rm{T}}
    \label{eq4}
\end{equation}
\( S^{s}\) then pass through a nonlinear activation function \( g \) 
to obtain the output \( U^{s}_{l+1} = g(S^{s}) \).
Due to this specific capsule tensor shape \((B, n, d)\), 
sparse attention routing can strictly adhere to the principles of dynamic routing 
to calculate similarity between capsule vectors for determining coupling coefficients.
The sparsity in routing preserves the richness of features 
and aligns with a parse-tree-like structure. 

\paragraph{Axial attention}
Sparse attention routing assigns coupling coefficient \( c_{ij} \) 
to each lower-level capsule \( u_{l,i} \), 
uniformly scaling all dimensions (i.e. instantiation parameter) 
within \( u_{l,i} \) by the same constant \( c_{ij} \). 
However, as not all instantiation parameter contribute equally to routing, 
the weight distribution for instantiation parameters within each capsule is needed. 
Nevertheless, traditional attention mechanisms based on dot products between vectors 
fail to account for dependencies between individual elements within capsules.

Axial attention decomposes a 3D input into capsule-wise ($n$-axis) and dimension-wise ($d$-axis) components, 
then applying attention independently to each axis \cite{ho2019axial,wang2020axial,valanarasu2021medical}. 
We focus attention along $d$-axis,
targeting the instantiation parameters in each capsule. 
Following \cite{ho2019axial}, unrelated axis $n$ is rearranged into the batch dimension $B$. 
Other process is similar to sparse attention, detailed in Alg. \ref{alg1}.
The shape of each tensor of axial attention is shown in Fig. \ref{SAA} with blue.
The outputs from both attentions, \( U^s_{l+1} \) and \( U^a_{l+1} \), 
are combined to get parent capsules $U_{l+1}$,
with both attentions computed in parallel to optimize training time.

\paragraph{Complexity analysis}
For \( U \in \mathbb{R}^{(B, n, d)} \), sparse attention has a computational complexity of \( O(n^2) \), 
while axial attention along d-axis incurs \( O(d) \). 
Given \( O(d) < O(n^2) \), the overall complexity remains \( O(n^2) \).
Compared with traditional attention routing where $\hat U \in \mathbb{R}^{(B, n, W \times H \times d)}$ 
with complexity of \( O( W^2 H^2 d^2) \) \cite{pucci2021self},
$\hat U \in \mathbb{R}^{(B, n_{l+1} ,n_l, d)}$ with \( O( n_l^2 d^2) \) \cite{mazzia2021efficient},
and $\hat U \in \mathbb{R}^{(B, w, h, n, d)}$ with Conv3D complexity of \( O( W H N D^2) \) \cite{choi2019attention},
SAA routing reduces computational costs.

\begin{algorithm}
    \centering
    \caption{SAA Routing}
    \label{alg1}
    \begin{algorithmic}[1]
    \Require $U_l \in \mathbb{R}^{(B, n_l, d_l)}$
    \Ensure $U_{l+1} \in \mathbb{R}^{(B, n_{l+1}, d_{l+1})}$

    \noindent \textbf{Compute sparse attention:}
    \State $\hat{U}_{l+1}^{s}\leftarrow \text{PConvCaps}(U_l)$
    \State $K^s, V^s \leftarrow \text{PointwiseConv}(U_l)$
    \State $C^s \leftarrow \alpha\text{-entmax}( \frac {\hat U_{l+1}^{s} {(K^{s})}^{\rm{T}}}{\sqrt{d_{l+1}}})$
    \State $S^s \leftarrow C^s {(V^s)}^{\rm{T}}$
    \State $U^s_{l+1} \leftarrow g(S^s)$
    
    \noindent \textbf{Axial attention along $d$-axis:}
    \State $\hat{U}_{l+1}^{a}\leftarrow Reshape (B \times n_{l+1},d_{l+1})$
    \State $K^a, V^a \leftarrow \text{PointwiseConv}(U_l)$
    \State $C^a \leftarrow \alpha\text{-Entmax}( \frac {\hat{U}_{l+1}^{a} {(K^{a})}^{\rm{T}}}{\sqrt{n_{l+1}}})$
    \State $S^a \leftarrow C^a  {(V^a)}^{\rm{T}}$
    \State $U^a_{l+1} \leftarrow g(S^a)$
    
    \State \Return $U_{l+1} \leftarrow U^s_{l+1} + U^a_{l+1}$
    \end{algorithmic}
\end{algorithm}
\vspace*{-10pt}

\subsection{Loss function} \label{3.3}

The loss function ensures that each concept capsule represents a specific concept, 
linking visual features in the image to the textual characteristics of the concept. 
It accommodates scenarios with and without concept ground truth labels. 

\subsubsection{When concept labels are available} \label{3.3.1}

\begin{table*}
  \vspace*{-10pt}
  \small
  \centering
  \caption{Comparison of ParseCaps with other models on the CE-MRI dataset. 
  Avg. is a macro average. 
  CapsNets variants are listed above the line, while CNN-based models are below it.}
  \label{table1.1}
  \vspace*{-5pt}
  \setlength{\tabcolsep}{1mm} 
    \begin{tabular}{cccccccc}
        \toprule
        Models & ACC$\uparrow$ & Avg. Precision$\uparrow$ & Avg. Recall$\uparrow$ & Avg. F1 Score$\uparrow$ & Avg. Specificity$\uparrow$ \\ 
        \midrule
        ParseCaps                 & \textbf{99.38} & \textbf{98.77} & \textbf{98.69} & \textbf{98.57} & \textbf{99.33} \\
        BrainCaps\cite{vimal2020effect}                 & 92.60 & 92.67 & 94.67 & 93.33 & -\\
        OrthCaps \cite{geng2024orthcaps}       & 92.57 & 86.18 & 86.27 & 85.84 & 94.25\\
        DeepCaps \cite{rajasegaran2019deepcaps}& 93.69 & 93.95 & 93.60 & 93.58 & 97.49\\
        CapsNet \cite{sabour2017dynamic}       & 92.16 & 93.36 & 91.47 & 92.08 & 93.84\\
        \midrule
        Modified GoogleNet \cite{sekhar2021brain}    & 94.90          & 94.76 & 93.69 & 94.30 & 97.22\\
        Block-wise VGG19 \cite{swati2019brain}     & 94.82          & 89.52 & 94.25 & 91.73 & 94.71\\
        \bottomrule
     \end{tabular}
  \vspace*{-10pt}
\end{table*}

\paragraph{Presentation loss}
To ensure the $i$-th concept capsule $u_{c,i}$ uniquely signifies the concept $p_i$,
the $u_{c,i}$ should be activate when $p_i$ is present in the image.
Let $Z = (z_1, z_2, \dots, z_m)$ represent the indicator for concept labels, 
where $z_i = 1$ indicates the presence of concept \( p_i \), and $z_i = 0$ otherwise. 
Accordingly, the presentation loss $L_p$ is defined in Eq. \ref{eq3.1}:
\begin{equation}
    L_p = \sum_{i=1}^{m} z_i \max(0, t^+_p - ||u_{c,i}||)^2 \\
    + (1 - z_i) \max(0, ||u_{c,i}|| - t^-_p)^2
    \label{eq3.1}
\end{equation}
Inspired by \cite{sabour2017dynamic}, activation margins $t^+_p = 0.9$ and $t^-_p = 0.1$ 
are set to ensure the length of the capsule vector $||u_{c,i}||$ exceeds $t^+_p$ for the correct concept, 
and remains below $t^-_p$ for the incorrect concept.

\paragraph{Triplet loss}
To link specific image regions with corresponding conceptual phrases, 
we use a linear embedding layer to map the instantiation capsule $u_{c,i}$ and concept $p_i$ 
into a joint latent space,
creating $\tilde u_{c,i}$ and $\tilde p_i$.
We want the embedding of the \(i\)-th instantiation capsule \(\tilde{u}_{c,i}\) corresponds to the \(i\)-th concept \(p_i\), 
while the embedding of the \(j\)-th capsule \(\tilde{u}_{c,j}\) does not, distinguishing the different embeddings.
The triplet loss $L_p$ ensures that the distance between $\tilde u_{c,i}$ and $\tilde p_i$ is less than 
the distance between $\tilde u_{c,j}$ and $\tilde p_i$ by a margin $t_t$.
The triplet loss $L_t$ is defined in Eq. \ref{eq3.2}:
\begin{equation}
    L_t = \sum_{i=1}^{m} \sum_{j=1, j \neq i}^{m} \max(0, \|\tilde{u}_{c,j}\\ 
    - \tilde{p}_i\|_2^2 - \|\tilde{u}_{c,i} - \tilde{p}_i\|_2^2 + t_t)
    \label{eq3.2}
\end{equation}

\paragraph{Overall loss}
Overall loss function $L$ is defined in Eq. \ref{eq3.3}:
\begin{equation}
    L = L_c+ \lambda L_p + \eta L_t
    \label{eq3.3}
\end{equation}
$L_c$ is the classification loss, which is the cross-entropy loss.
$\lambda$ and $\eta$ are the weight hyperparameters.

\subsubsection{When concept labels are unavailable} \label{3.3.2}

When explicit concept labels are unavailable, \( L_p \) and \( L_t \) cannot be used. 
Nevertheless, the parse-tree-like structure, where each node represents a specific image region or component, 
aids the model in understanding the relationships and hierarchy among different image parts.  
Additionally, the instantiation parameter can inherently signify the presence of specific concepts. 
Therefore, it is adequate to motivate the model to learn concepts that reflect the semantics of the input image \( x_i \).
To measure reconstruction error, the reconstruction loss \( L_r \) is added to the overall loss \( L \).
If the model's conceptual understanding is insufficient for an accurate reconstruction of \( x_i \), 
the \( L_r \) penalizes the model. 
We utilize an $L_2$ loss for this purpose and assign a weight of \( \gamma \).
All hyperparameters are detailed in supplementary material \ref{hyperparameters}.
The overall loss function $L$ is defined as Eq. \ref{eq3.4}:
\begin{equation}
  L = L_c + \gamma L_r
  \label{eq3.4}
\end{equation}

\section{Experiments}

\subsection{Experimental setup} \label{4.1}

\paragraph{Setup and datasets} 
ParseCaps was developed using PyTorch 12.1 and Python 3.9, accelerated by eight GTX-3090 GPUs. 
We set the learning rate to 2.5e-3, batch size to 64, and weight decay to 5e-4. 
The model was trained for 300 epochs using the AdamW optimizer, 
and a 5-cycle linear warm-up.
We evaluated ParseCaps with Contrast Enhanced Magnetic Resonance Images (CE-MRI) \cite{Cheng2017}, 
PH$^2$ \cite{6610779} and Derm7pt (D7) \cite{Kawahara2019-7pt} datasets.
CE-MRI has 3064 MRI T1w post GBCA images from
233 patients.
PH$^2$ and D7 are two skin diagnostic datasets with 200 and 2000 images, respectively. 

\paragraph{Concept label acquisition} 
By dividing textual descriptions using a concept parser, 
we get conceptual phrases composed of a noun and corresponding adjectives, 
serving as concept-level annotations \cite{wickramanayake2021comprehensible}. 
Based on this rule 
and the annotations in the dermoscopic standards \cite{patricio2023coherent}, 
we select ``Atypical Pigment Network" (APN), 
``Typical Pigment Network" (TPN), 
``Blue Whitish-Veil" (BWV), 
``Irregular Streaks" (ISTR), 
``Regular Streaks" (RSTR), 
``Regular Dots and Globules" (RDG), 
and ``Irregular Dots and Globules" (IDG) as the concept labels for PH$^2$ and D7.

\begin{table*}
    \vspace*{-10pt}
    \small
    \centering
    \caption{Performance of each class on CE-MRI. P, R, S, and F1 are precision, recall, specificity, and F1 score, respectively.}
    \vspace{-5pt}
    \label{table1.2}
    \setlength{\tabcolsep}{1mm}
      \begin{tabular}{c|c|c|c|c|c|c|c|c|c|c|c|c}
           \toprule
           Classes & \multicolumn{4}{c|}{Meningioma} & \multicolumn{4}{c|}{Glioma} & \multicolumn{4}{c|}{Pituitary tumor} \\ 
           \midrule
           Metrics & P$\uparrow$ & R$\uparrow$  & S$\uparrow$ & F1 $\uparrow$ & P & R  & S & F1 & P & R  & S & F1\\
           \midrule
           ParseCaps                                & \textbf{96.92} & \textbf{95.64} & \textbf{99.79} & \textbf{96.12} & \textbf{99.74} & \textbf{99.39} & \textbf{99.63} & \textbf{99.54} & \textbf{98.08} & \textbf{99.06} & \textbf{99.39} & \textbf{98.89}\\
           OrthCaps \cite{geng2024orthcaps}         & 89.14 & 83.15 & 97.84 & 85.37 & 96.35 & 97.66 & 91.77 & 97.03 & 91.13 & 96.19 & 96.83 & 96.40\\
           DeepCaps \cite{rajasegaran2019deepcaps}  & 94.71 & 91.79 & 98.47 & 91.04 & 98.33 & 97.12 & 95.41 & 96.71 & 90.79 & 96.10 & 95.22 & 95.14\\
           CapsNet \cite{sabour2017dynamic}         & 88.57 & 94.33 & 86.33 & 91.06 & 86.78 & 88.09 & 87.70 & 92.36 & 91.10 & 92.54 & 86.05 & 95.14\\
           BrainCaps \cite{vimal2020effect}         & 85.00 & 94.00 & -     & 89.00 & 98.00 & 96.00 & -     & 97.00 & 95.00 & 94.00 & -     & 94.00\\
           Modified GoogleNet \cite{sekhar2021brain}    & 93.78 & 86.98 & 98.19 & 90.25 & 96.02 & 97.00 & 96.00 & 96.51 & 94.48 & 97.10 & 97.47 & 95.77\\
           Block-wise VGG19 \cite{swati2019brain}       & 87.97 & 89.98 & 96.42 & 88.88 & 93.26 & 95.97 & 93.79 & 94.52 & 87.34 & 96.81 & 93.93 & 91.80\\
          
           \bottomrule
      \end{tabular}
    \vspace*{-10pt}
  \end{table*}

\subsection{Classification performance comparison} \label{4.2}

We choose CE-MRI dataset to evaluate the classification ability of ParseCaps in Tab. \ref{table1.1}, 
because it is a typical dataset with comparable CapsNets varients.
ParseCaps outperforms other models in accuracy (ACC), with a notable score of 99.38\%.
ParseCaps also leads in average precision, recall, F1 score, and specificity, 
underscoring its robustness and effectiveness in handling CE-MRI dataset.
Tab. \ref{table1.2} presents class-specific metrics for 
Meningioma, Glioma, and Pituitary tumor,  
where ParseCaps shows high performance across all metrics, 
notably achieving a specificity of 99.79\% for Meningioma  
and a precision of 99.74\% for Glioma. 
This highlights ParseCaps' superior sensitivity and precision in distinguishing different brain tumor types.
Tab. \ref{table1.3} shows ParseCaps' performance on skin datasets, including PH$^2$, D7, and a combined PH$^2$D7. 
ParseCaps has the highest scores on two datasets, achieving 97.53\% and 87.96\% respectively.
Although the performance on D7 is lower than the coherent CNN model, it is still competitive considering the interpretability. 
Overall, these results affirm ParseCaps' efficacy across various medical image tasks,  
with additional classification results in the supplementary material \ref{5.4}.

\begin{table}[t]
    \caption{Classification accuracy performance on skin datasets. 
    PH$^2$D7 is a combination of the PH$^2$ and D7 datasets created by \cite{patricio2023coherent}.}
    \label{table1.3}
    \vspace*{-5pt}
    \resizebox{\linewidth}{!}{
      \begin{tabular}{cccc}
        \toprule
        Models & PH$^2$$\uparrow$ & D7$\uparrow$ & PH$^2$D7$\uparrow$  \\ 
        \midrule
        ParseCaps &  \textbf{97.53} & 83.42 & \textbf{87.96} \\
        Coherent CNN \cite{patricio2023coherent} & 96.00  & 84.06 & 84.44  \\
        Skin VGGNet \cite{lopez2017skin} & 90.67  & 76.15 & 79.23  \\
        CCNN \cite{wickramanayake2021comprehensible} & 93.33  & \textbf{84.27} & 83.96  \\
        SqueezeNet \cite{abayomi2021malignant} & 92.18 & - & - \\
        \bottomrule
   \end{tabular}
    }
    \vspace*{-8pt}
\end{table}

\subsection{Ablation study} \label{4.3}

\subsubsection{PConvCaps layer} \label{4.3.1}

ParseCaps, DeepCaps, and OrthCaps all have a ConvCaps Block. 
Tab. \ref{table2} compares the capsule counts ($n$) and dimensions ($d$) within these models 
to highlight PConvCaps layer's role in the parse-tree-like structure. 
Unlike DeepCaps and OrthCaps, which generally increase the number of capsules while maintaining or reducing their dimensions as layers deepen, 
ParseCaps consistently reduces the number of capsules and increases their dimensions, aligning with the parse-tree-like structure. 
Furthermore, ParseCaps adaptively adjusts \( n \) and \( d \) according to the dataset's complexity.
For example, as shown in the left column of ParseCaps, larger and more complex images of CE-MRI lead to correspondingly larger \( n \) and \( d \),  
increasing the amount of information the capsules can represent. 
For simpler datasets like MNIST \cite{726791} in the right column, \( n \) and \( d \) are minimized to simplify the model and prevent overfitting, 
with automatic adjustments based on input image size that require no extra operation.

\begin{table}[t]
  \caption{Comparison of the number of capsules \( n \) and the dimensions \( d \). 
  ParseCaps is tested under the CE-MRI and MNIST, while others are tested under CE-MRI.
  }
  \label{table2}
  \vspace*{-5pt}

  \resizebox{\linewidth}{!}{
    \begin{tabular}{c|c|c|c|c}
         \toprule
          Models & \multicolumn{2}{c|}{ParseCaps} & DeepCaps & OrthCaps \\ 
         \midrule
          Metrics      & $n$ | $d$ & $n$ | $d$  & $n$ | $d$ & $n$ | $d$ \\
         \midrule
         Primary Capsule Layer & 784 | 8 & 14 | 2 & 1 | 128 & 16 | 16 \\
         ConvCaps Block 1      & 196 | 16 & 7 | 4 & 4 | 32   & 32 | 16\\
         ConvCaps Block 2      & 49 | 36 & 4 | 8 & 8 | 32   & 64 | 16\\
         ConvCaps Block 3      & 13 | 64 & 2 | 16 & 8 | 32   & 128 | 16\\
         FullyCaps Layer       & 1 | 64 & 1 | 16 & 10 | 16  & 10 | 16\\
         \bottomrule
    \end{tabular}
 }

  \vspace*{-10pt}
\end{table}

\subsubsection{Loss functions} \label{4.3.2}

We assess the impact of each loss using Explanation Error (EE) \cite{sarkar2022framework} and classification accuracy (ACC), 
measured EE as the $L_2$ distance, 
where a lower EE indicates better alignment with ground truth concepts.
Tab. \ref{table5} shows that 
the combination of the classification loss \( L_c \), the presentation loss \( L_p \) and the triplet loss \( L_t \)
achieves the best performance, with ACC of 97.53\% and EE of 0.98.
Removing \( L_t \) leads to a drop in performance, with ACC falling to 97.25\% and EE increasing to 1.13, 
underscoring \( L_t \)'s critical role in linking concept features with visual features.
The ablation study of the reconstruct loss $L_r$ is in the supplementary material \ref{5.3}.

\subsubsection{Attention routing} \label{4.3.3}

We test the performance of routing algorithms in a basic CapsNet model,  
featuring a convolutional layer with a kernel size of 3 and stride of 2, a primary capsule layer, and a digit capsule layer, 
with the routing between the last two layers. 
As shown in Tab. \ref{table7}, 
SAA routing outperforms others with an accuracy of 99.01\% and performs best in FPS and FLOPS,
showcasing superior efficiency.
Although dynamic routing records a decent accuracy of 98.35\%, 
it suffers from high computational costs (40609M FLOPS) and the lowest FPS (252.78).

\begin{table}[t]
  \small
  \centering
  \caption{Ablation study on loss functions. Results are tested with ParseCaps on PH$^2$ dataset.}
  \vspace{-5pt}
  \label{table5}

    \resizebox{0.5\linewidth}{!}{
      \begin{tabular}{ccc}
        \toprule
        Method           & ACC $\uparrow$ & EE $\downarrow$ \\ 
        \midrule
        $L_c$        & 94.89 & 4.02 \\
        $L_c + L_p$  & 95.01 & 3.93 \\
        $L_c + L_t$  & 97.25 & 1.13 \\
        $L_c + L_f + L_p$  & \textbf{97.53} & \textbf{0.98} \\
        \bottomrule
      \end{tabular}
    }
  
\end{table}

\begin{table}[t]
  \small
  \centering
  \vspace{-2pt}
  \caption{Ablation study on different routing algorithms.
  Results are tested on a 3-layer CapsNet model, which is trained on the MNIST dataset for 100 epochs.}
  \vspace{-5pt}
  \label{table7}
    \resizebox{0.7\linewidth}{!}{
      \centering
      \begin{tabular}{cccc}
        \toprule
        Routing           & FPS $\uparrow$ & FLOPS $\downarrow$ & ACC $\uparrow$\\ 
        \midrule
        SAA       & \textbf{387.88} & \textbf{279M} & \textbf{98.66} \\
        Attention  & 277.55 & 10627M & 98.23 \\
        Dynamic    & 252.78 & 40609M & 98.35 \\
        \bottomrule
      \end{tabular}
    }
    \vspace{-15pt}
  
\end{table}

\subsection{Effectiveness of parse-tree-like structure} \label{4.5}


\subsubsection{Robustness analysis} \label{4.5.3}

In the parse-tree-like structure, lower-level capsules detect basic features like edges and corners, 
and higher-level capsules aggregate these basic features to represent more complex entities. 
This bottom-up feature integration enables the model to maintain consistent recognition under affine transformations, thus enhancing robustness.
Following the experimental protocol of \cite{sabour2017dynamic}, we train ParseCaps for 100 epochs on the MNIST 
and test it on the affNIST dataset \cite{tieleman2013affnist}, 
which subjects images to random affine transformations like rotations, scaling, and translations.
We use a baseline capsule network with SAA routing but without the parse-tree-like structure.
Tab. \ref{table6} shows that ParseCaps demonstrates superior accuracy on affNIST,  
achieving 84.32\%, compared to baseline's 79.00\% and CNN's 66.00\%, confirming the parse-tree-like structure's role in boosting model robustness.

\begin{table}[t]
  \small
  \vspace*{-5pt}
  \caption{ Robustness analysis on the affNIST dataset. MNIST and affNIST mean the accuracies on these datasets, respectively.
  }
  \label{table6}
  \vspace*{-10pt}
  \centering
  \resizebox{0.5\linewidth}{!}{
  \setlength{\tabcolsep}{1mm}
    \begin{tabular}{ccc}
        \toprule
        Variants           & MNIST $\uparrow$ & affNIST $\uparrow$ \\ 
        \midrule
        ParseCaps        & \textbf{99.30} & \textbf{84.32}  \\
        Baseline         & 99.23 & 79.00 \\
        CNN              & 99.22 & 66.00 \\
        \bottomrule
      \end{tabular}
  }
\end{table}

\subsubsection{Redundancy analysis} \label{4.5.2}

We use capsule similarity distributions \cite{geng2024orthcaps} to compare the redundancy of the top-layer capsules among different capsule network variants, which is detailed in the supplementary material \ref{5.5}.

\subsubsection{Concept interpretability evaluation} \label{4.5.1}

\begin{figure}[t]
    \centering
    \includegraphics[width=0.9\columnwidth]{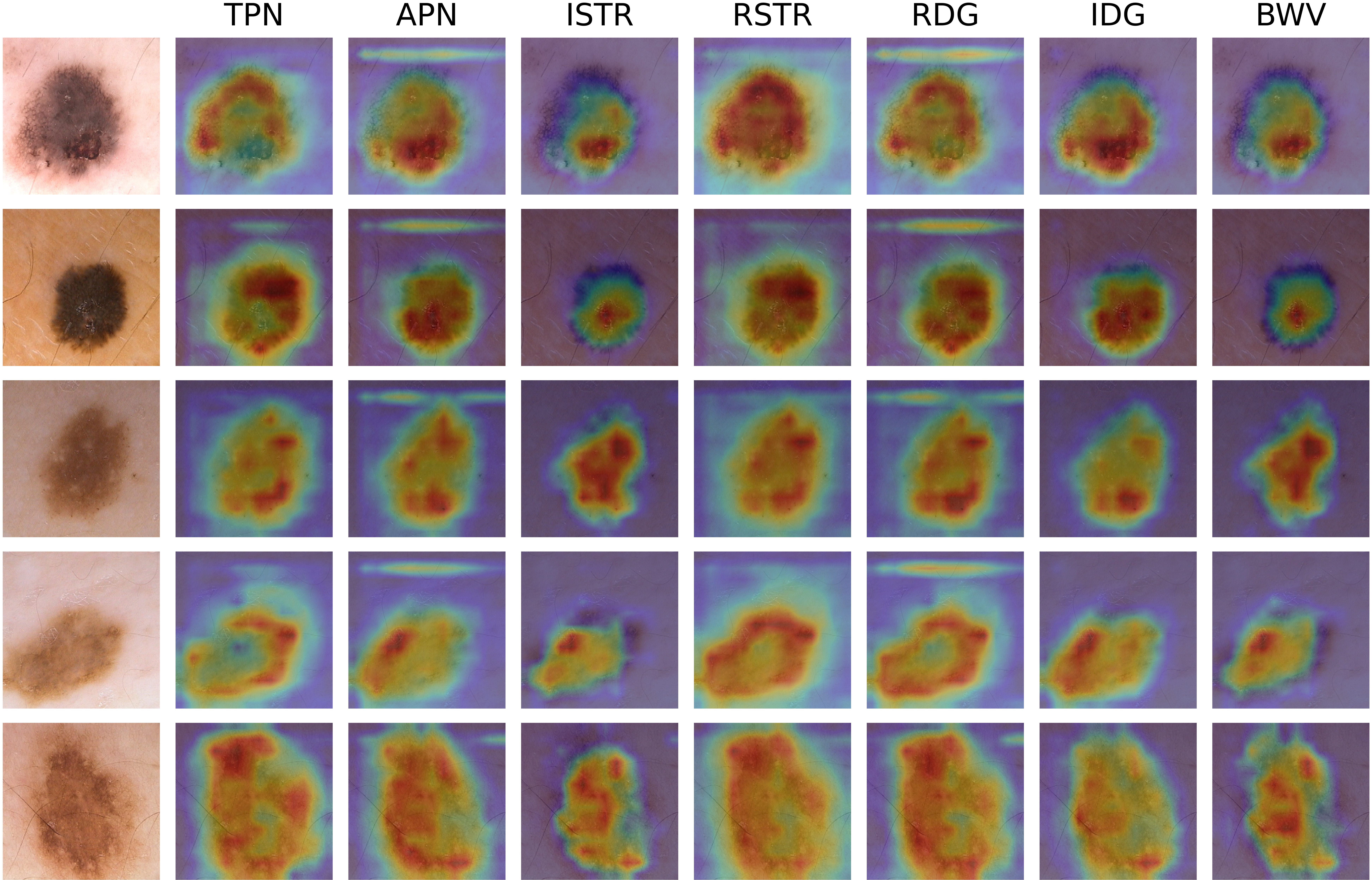} 
    \caption{Visualization of the concept capsules with ground truth in ParseCaps.}
    \vspace*{-12pt}
    \label{Fcon_A}
\end{figure}

\begin{figure}[t]
    \vspace*{-7pt}
    \centering
    \includegraphics[width=0.9\columnwidth]{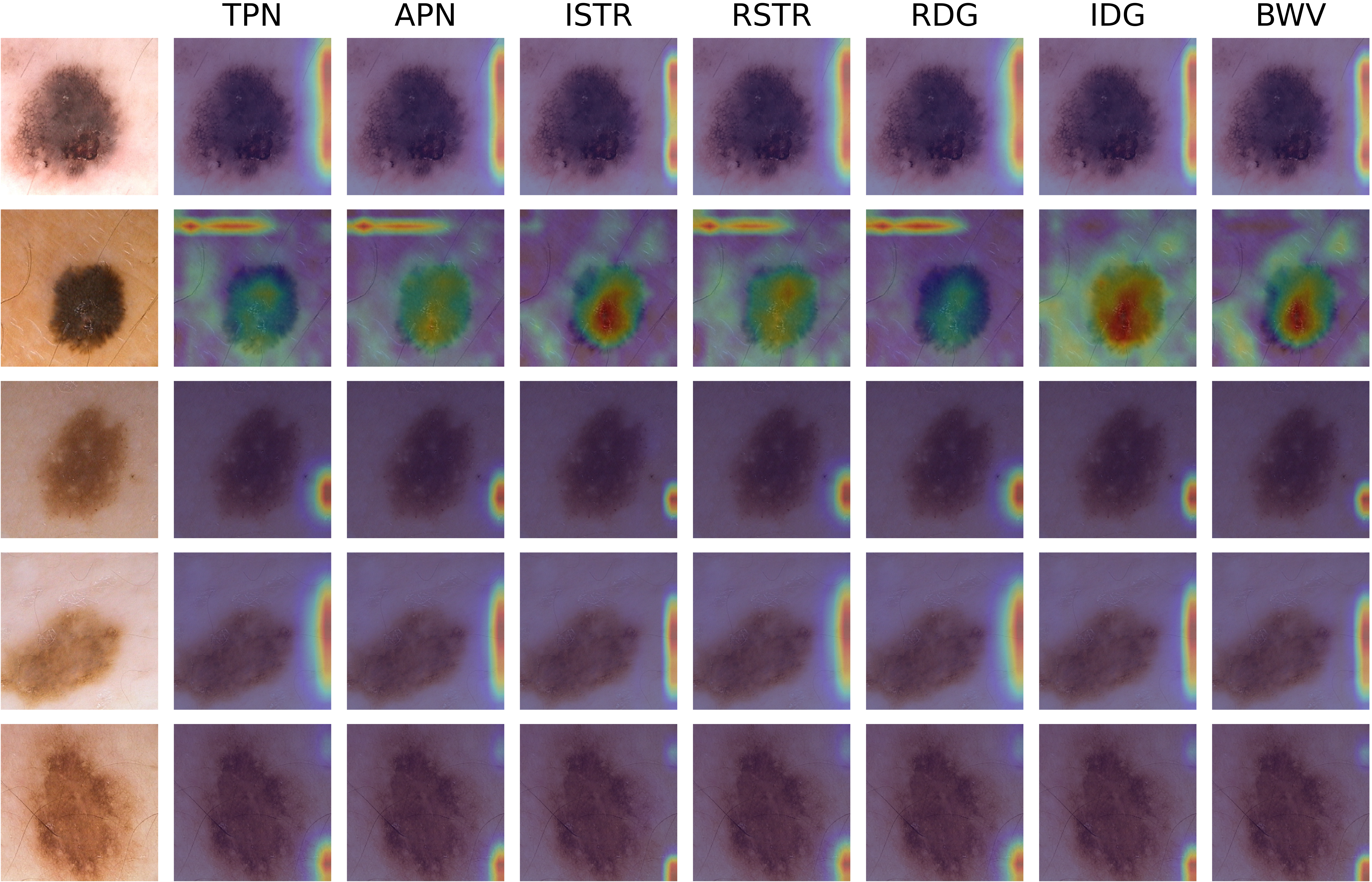} 
    \caption{Visualization of the concept capsules in the baseline model.}
    \vspace*{-10pt}
    \label{Fcon_B}
\end{figure}

To validate the interpretability of ParseCaps, we analyze the concept capsules within the model through visualization, 
as shown in Fig. \ref{Fcon_A}.
The highlighted areas coincide with the lesion areas, proving the detection of key regions for melanoma.
According to the ABCD rule \cite{nachbar1994abcd},
these areas clinically correspond to common signs of melanoma: Asymmetry, irregular Borders, multiple Colors, and Dermoscopic structures \cite{patricio2023coherent}.
We construct a baseline model with the same structure as ParseCaps except the parse-tree-like structure, directly connecting all capsules to the concept layer.
As shown in Fig. \ref{Fcon_B}, its concept capsules fail to capture critical areas and instead focus on edge features, 
which should be captured by lower-level capsules, validating the effectiveness of parse-tree-like structure.

\begin{figure}
    \centering
    \includegraphics[width=0.9\columnwidth]{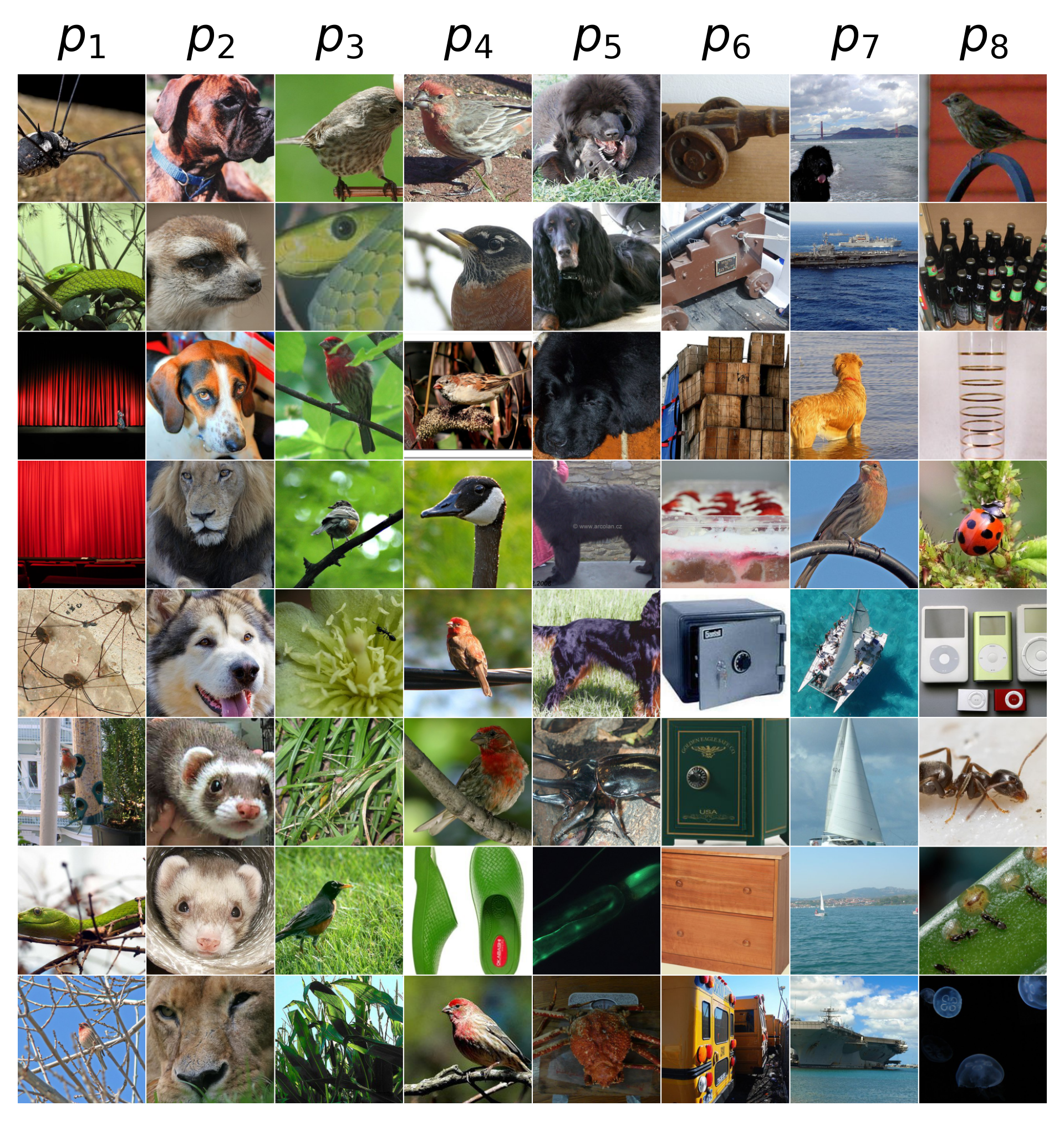} 
    \vspace*{-3pt}
    \caption{Visualization of the concept capsules without ground truth on ImageNet-mini dataset.}
    \label{unsupervised}
    \vspace*{-10pt}
\end{figure}

When concept ground truth is unavailable, ParseCaps leverages internal entity relationships to provide interpretable concepts within images.
We choose prototypes that maximize the activation values of each concept capsule to define its specific meanings. 
Due to the difficulty non-medical experts face in annotating the medical concepts of the selected prototypes, 
we have chosen to test ParseCaps on ImageNet-mini \cite{NIPS2016_90e13578}, 
which has more images and concepts that are easier to recognize.
As depicted in Fig. \ref{unsupervised}, 
\(p_1\) captures entities with vertical line features, 
\(p_2\) identifies an animal's face, 
\(p_3\) focuses on green entities, 
\(p_4\) represents a left-facing bird with a pointed beak, 
\(p_5\) encompasses black or dark entities, 
\(p_6\) captures cubic shapes, 
\(p_7\) is associated with the blue sky or sea, 
and \(p_8\) recognizes circular features. 
Although there are occasional misidentifications, such as slippers in \(p_4\), 
ParseCaps consistently demonstrates interpretable ability without concept supervision.

\section{Conclusion and limitations}

In this paper, we introduce ParseCaps, 
which incorporates a parse-tree-like structure and loss functions to build an interpretable capsule network. 
ParseCaps uses SAA routing to optimize the connections between child and parent capsules and reduce computational complexity,  
while PConvCaps layer is suitable for the parse-tree-like structure.
ParseCaps advances the interpretability of deep CapsNets in medical classification.
However, ParseCaps has limitations. 
First, the parse-tree-like structure is not a strict parse tree with single-parent connections.
Second, due to limited medical knowledge, 
identifying conceptual meanings for prototypes is challenging. 
Currently, ParseCaps lacks experiments in interpretability visualization without concept ground truth on medical datasets. 
We hope for continued exploration of unsupervised interpretability on medical datasets.




\newpage
\bibliography{aaai25}

\newcommand{\answerTODO}[1]{\textbf{#1}} 

\newpage
\newpage

\section{Reproducibility checklist}

\begin{enumerate}
  \item This paper:
    \begin{enumerate}
      \item Includes a conceptual outline and/or pseudocode description of AI methods introduced (yes/partial/no/NA)
      \answerTODO{Yes}
      \item Clearly delineates statements that are opinions, hypothesis, and speculation from objective facts and results (yes/no)
      \answerTODO{Yes}
      \item Provides well marked pedagogical references for less-familiare readers to gain background necessary to replicate the paper (yes/no)
      \answerTODO{Yes}
    \end{enumerate}
  \item Does this paper make theoretical contributions? (yes/no) \answerTODO{Yes}
  \\ If yes, please complete the list below.
    \begin{enumerate}
      \item All assumptions and restrictions are stated clearly and formally. (yes/partial/no)
      \answerTODO{Yes}
      \item All novel claims are stated formally (e.g., in theorem statements). (yes/partial/no)
      \answerTODO{Yes}
      \item Proofs of all novel claims are included. (yes/partial/no)
      \answerTODO{Yes}
      \item Proof sketches or intuitions are given for complex and/or novel results. (yes/partial/no)
      \answerTODO{Yes}
      \item Appropriate citations to theoretical tools used are given. (yes/partial/no)
      \answerTODO{Yes}
      \item All theoretical claims are demonstrated empirically to hold. (yes/partial/no/NA)
      \answerTODO{Yes}
      \item All experimental code used to eliminate or disprove claims is included. (yes/no/NA)
      \answerTODO{Yes}
    \end{enumerate}
  \item Does this paper rely on one or more datasets? (yes/no) \answerTODO{Yes}
  \\ If yes, please complete the list below.
    \begin{enumerate}
      \item A motivation is given for why the experiments are conducted on the selected datasets (yes/partial/no/NA)
      \answerTODO{Yes}
      \item All novel datasets introduced in this paper are included in a data appendix. (yes/partial/no/NA)
      \answerTODO{Yes}
      \item All novel datasets introduced in this paper will be made publicly available upon publication of the paper with a license that allows free usage for research purposes. (yes/partial/no/NA)
      \answerTODO{Yes}
      \item All datasets drawn from the existing literature (potentially including authors’ own previously published work) are accompanied by appropriate citations. (yes/no/NA)
      \answerTODO{Yes}
      \item All datasets drawn from the existing literature (potentially including authors’ own previously published work) are publicly available. (yes/partial/no/NA)
      \answerTODO{Yes}
      \item All datasets that are not publicly available are described in detail, with explanation why publicly available alternatives are not scientifically satisficing. (yes/partial/no/NA)
      \answerTODO{Yes}
    \end{enumerate}
  \item Does this paper include computational experiments? (yes/no) \answerTODO{Yes}
  \\ If yes, please complete the list below.
    \begin{enumerate}
      \item Any code required for pre-processing data is included in the appendix. (yes/partial/no).
      \answerTODO{Yes}
      \item All source code required for conducting and analyzing the experiments is included in a code appendix. (yes/partial/no)
      \answerTODO{Yes}
      \item All source code required for conducting and analyzing the experiments will be made publicly available upon publication of the paper with a license that allows free usage for research purposes. (yes/partial/no)
      \answerTODO{Yes}
      \item All source code implementing new methods have comments detailing the implementation, with references to the paper where each step comes from (yes/partial/no)
      \answerTODO{Yes}
      \item If an algorithm depends on randomness, then the method used for setting seeds is described in a way sufficient to allow replication of results. (yes/partial/no/NA)
      \answerTODO{Yes}
      \item This paper specifies the computing infrastructure used for running experiments (hardware and software), including GPU/CPU models; amount of memory; operating system; names and versions of relevant software libraries and frameworks. (yes/partial/no)
      \answerTODO{Yes}
      \item This paper formally describes evaluation metrics used and explains the motivation for choosing these metrics. (yes/partial/no)
      \answerTODO{Yes}
      \item This paper states the number of algorithm runs used to compute each reported result. (yes/no)
      \answerTODO{Yes}
      \item Analysis of experiments goes beyond single-dimensional summaries of performance (e.g., average; median) to include measures of variation, confidence, or other distributional information. (yes/no)
      \answerTODO{Yes}
      \item The significance of any improvement or decrease in performance is judged using appropriate statistical tests (e.g., Wilcoxon signed-rank). (yes/partial/no)
      \answerTODO{Yes}
      \item This paper lists all final (hyper-)parameters used for each model/algorithm in the paper’s experiments. (yes/partial/no/NA)
      \answerTODO{Yes}
      \item This paper states the number and range of values tried per (hyper-) parameter during development of the paper, along with the criterion used for selecting the final parameter setting. (yes/partial/no/NA)
      \answerTODO{Yes}
    \end{enumerate}
\end{enumerate}


\newpage

\appendix

\section{Appendix / supplemental material}

\subsection{Symbols and abbreviation used in this paper}
\label{5.1}

In the following notation, a capsule is represented as a vector, hence denoted by lowercase letters. 
A collection of capsules forming a matrix is represented by uppercase letters.

\begin{table}[H] 
   \centering
    \caption{Symbols and abbreviations used in this paper.}
    \resizebox{\columnwidth}{!}{
      \begin{tabular}{ll}
        \toprule
        Symbol & Description  \\ 
        \midrule
        $x$ & Input image \\
        $\Phi^0$ & Features from the initial convolutional block \\
        $B$ & Batch size \\
        $f_{0}$ & Number of feature map in the initial convolutional block \\
        $W_{0}$ & Width of feature map in the initial convolutional block \\
        $H_{0}$ & Height of feature map in the initial convolutional block \\
        $U_1$ & Primary capsule layer \\
        $n_1$ & Capsule count in the primary capsule layer \\
        $d_1$ & Capsule dimension in the primary capsule layer \\
        $l$ & Layer index \\
        $\hat{U}_{l+1}$ &  Prediction of the layer $l+1$ capsules\\
        $u_{l,i}/(U_l)$ & Capsule $i$ / (matrix) at layer $l$ \\
        $c_{ij}$ & Coupling coefficient \\
        $u_{class,i}/(U_{class})$ & Capsule / (matrix) in the classification layer \\
        $n_{class}$ & The number of classes \\
        $d_{class}$ & Capsule dimension in the classification layer \\
        $U_{c}$ & The concept capsule matrix \\
        $m$ & Number of concept capsules \\
        $d_c$ & Capsule dimension in the concept capsule \\
        $u_{c,i}$ & The $i$-th concept capsule \\
        $u_{g}$ & The global capsule \\
        $p_i$ & The $i$-th concept label, a word phrase \\
        $w_l$ & the width of feature map / $\sqrt n$ in layer $l$ \\
        $d_l$ & Capsule dimension in layer $l$ \\
        $n_l$ & Capsule count in layer $l$ \\
        $C^{s}$ & Coupling coefficient matrix in sparse attention routing \\
        $\hat U_{l+1}^{s}$ & Prediction of the layer $l+1$ capsules (The capsule query in sparse attention routing) \\
        $K^s$ & Capsule key in sparse attention routing \\
        $\tau$ & self-adaption threshold of $\alpha$-Entmax \\
        $\alpha$ & a hyperparameter controlling the sparsity of the attention map \\
        $S^{s}$ & The votes in sparse attention routing \\
        $g$ & Activation function \\
        $U^{s}_{l+1}$ & The output of the sparse attention routing \\
        $\hat U_{l+1}^{a}$ & Prediction of the layer $l+1$ capsules (The capsule query in axial attention routing) \\
        $K^a$ & Capsule key in axial attention routing \\
        $C^{a}$ & Coupling coefficient matrix in axial attention routing \\
        $S^{a}$ & The votes in axial attention routing \\
        $U^{a}_{l+1}$ & The output of the axial attention routing \\
        $Z$ &  The indicator for concept labels \\
        $z_i$ & The $i$-th concept indicator \\
        $u_{c,i}$ & The $i$-th concept capsule \\
        $L_{p}, L_{c}, L_{t}, L_{r}, L$ & Presentation loss, Classification loss, Triplet loss, Reconstruction loss, Total loss \\
        $t^+_p, t^-_p$ & Margins of the presentation loss \\
        $\tilde u_{c,i}$ & The embedding of the concept capsule \\
        $\tilde p_i$ & The embedding of the concept label \\
        $t_{t}$ & Margins of the triplet loss \\
        $\lambda, \eta, \gamma$ & The weights of the loss functions \\
        
        
        \bottomrule
      \end{tabular}
    }
\end{table}

\subsection{Experimental setups}
\label{5.2}

\subsubsection{Hyperparameters}
\label{hyperparameters}

\begin{table}[H] 
   \centering
   \caption{Hyperparameters used in the experiments.}
   \resizebox{\columnwidth}{!}{
   \begin{tabular}{ll}
       \toprule
       Hyperparameter & Value  \\ 
       \midrule
       Batchsize & 64 (8 parallelled)  \\
       Learning rate & 2.5e-3 \\
       Weight decay & 5e-4 \\
       Optimizer & AdamW \\
       Scheduler & CosineAnnealingLR and 5-cycle linear warm-up \\
       Epochs & 300 \\
       Data augmentation & RandomHorizonFlip, RandonClip with padding of 4 \\
       Dropout & 0.25 \\
       $m^+$ & 0.9 \\
       $m^-$ & 0.1 \\
       $d$ & [16,36,64] \\
       $\lambda$ & 0.5 \\
       $\eta$ & 0.1 \\
       $\gamma$ & 0.0005 \\
       \bottomrule
       \end{tabular}
   }
\end{table}

\subsubsection{setups}

In this section, we describe the necessary Python library and corresponding version for the experiments in the main paper.

\begin{table}[H] 
   \centering
   \small
   \caption{Python library and version used in the experiments.}
   \begin{tabular}{ll}
       \toprule
       Library & Version  \\ 
       \midrule
       pytorch & 1.12.1  \\
       numpy & 1.24.3 \\
       opencv-python & 4.7.0.72 \\
       pandas & 2.0.2 \\
       pillow & 9.4.0 \\
       torchvision & 0.13.1 \\
       matplotlib & 3.7.1 \\
       icecream & 2.1.3 \\
       seaborn & 0.12.0 \\
       \bottomrule
       \end{tabular}
   
\end{table}


\subsection{Supplementary experiments of reconstruction loss}
\label{5.3}

Reconstruction loss penalizes the differences between reconstructed images and original images, 
aiding the model in learning the fundamental structures of the data, thereby enhancing the model's capabilities. 
Tab. \ref{table3} presents the performance of ParseCaps under different reconstruction loss weights ($\eta$). 
We tested commonly used $\eta$ for capsule network variants: 0.005, 0.001, and 0. 
The results show that the highest accuracy (ACC) of 99.375\% is achieved when $\eta$ is 0.005. 
ACC slightly decreases to 98.90\% when $\eta$ is 0.001. 
Removing the reconstruction loss results in a further drop in accuracy to 98.32\%, 
confirming the importance of reconstruction loss to model performance.

\begin{table}[h]
  \centering
  \caption{Ablation study on reconstruction loss. ACC is tested on CE-MRI.}
  \label{table3}
  \small
  \begin{tabular}{cccc}
      \toprule
      Model & $\eta$ & PARAM & ACC$\uparrow$  \\ 
      \midrule
      ParseCaps & 0.005  & 3.6M & \textbf{99.38}  \\
      ParseCaps & 0.001 & 3.6M & 98.90 \\
      ParseCaps w/o Reconstruction & 0 & 490K & 98.32 \\
      \bottomrule
  \end{tabular}
\end{table}

\subsubsection{Supplementary experiments of classification performance}
\label{5.4}

Tab. \ref{table0} presents additional classification comparison results on CE-MRI datasets.
The comparison includes ParseCaps and other CapsNets variants, which don't have available codes.
So we use the results in the paper.
These results further confirm ParseCaps' superior performance in medical image classification tasks.

\begin{table}[h]
  \centering
  \caption{Additional classification comparison on CE-MRI datasets.}
  \label{table0}
  \small
  \begin{tabular}{cc}
      \toprule
      Models & ACC \\ 
      \midrule
      ParseCaps & \textbf{99.38}  \\
      \cite{afshar2020bayescap} & 68.30 \\
      \cite{adu2019dilated}     & 95.54 \\
      \cite{afshar2018brain}    & 86.56 \\
      \bottomrule
  \end{tabular}
\end{table}

\subsection{Supplementary experiments of redundancy analysis}
\label{5.5}

\begin{figure}
  \vspace*{-20pt}
  \centering
  \includegraphics[width=0.75\columnwidth]{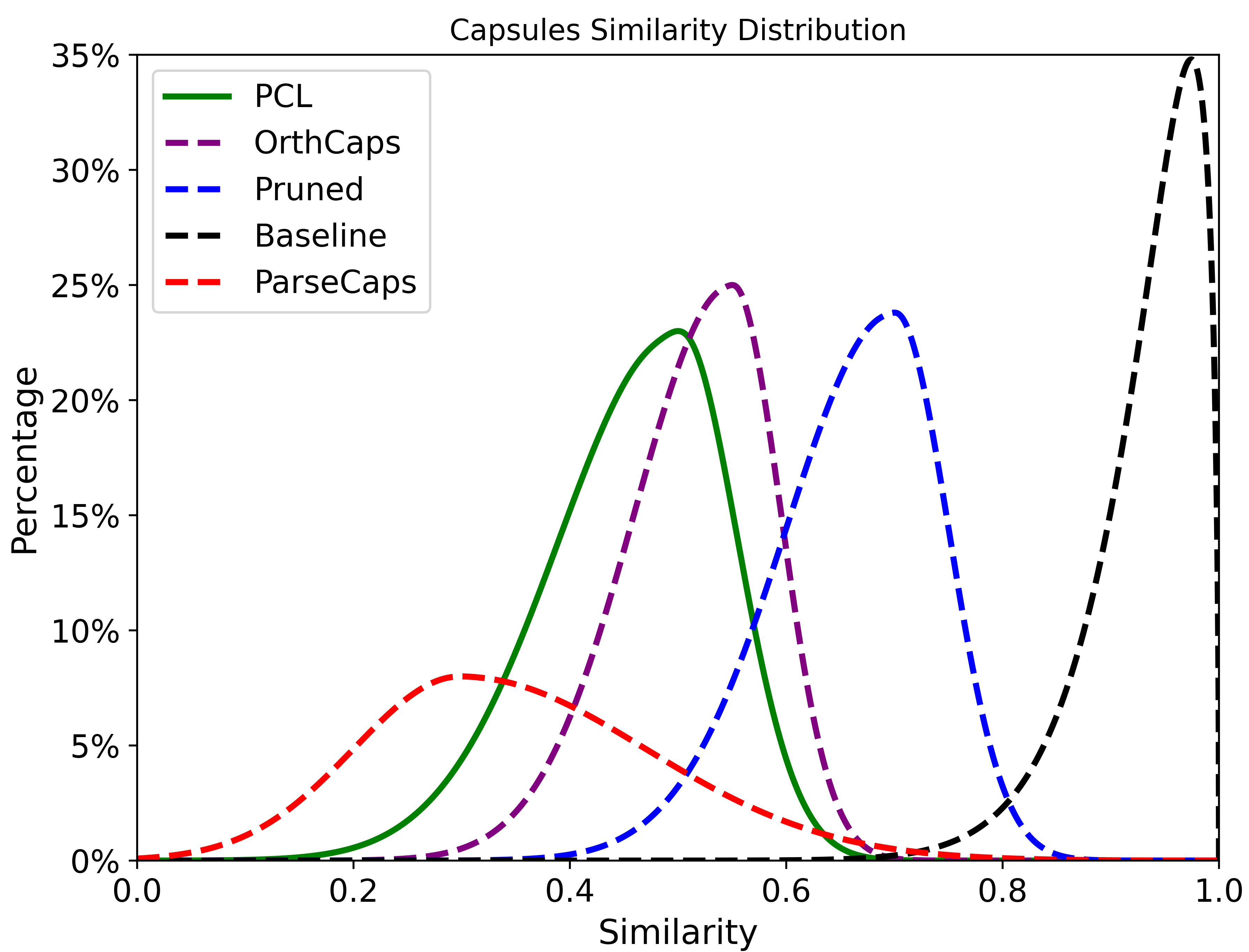} 
  \caption{Redundancy comparison among varients.
  The $x$-axis is capsule similarity and $y$-axis is capsule count percentage.
  PCL, Pruned and Baseline denote the primary capsule layer, \cite{geng2024orthcaps} and CapsNet, respectively.}
  \label{pruned}
  \vspace*{-8pt}
\end{figure}

The further left the curve, the lower the capsule redundancy.
ParseCaps and other capsule network variants is set to generate the same number of capsules in the primary capsule layer to ensure a fair comparison.
As shown in Fig. \ref{pruned}, ParseCaps exhibits lower redundancy compared to other capsule network variants, with fewer high-level capsules, 
and achieving information aggregation. 
This redundancy reduction is not achievable by other capsule network variants where the number of capsules gradually increases.

\subsection{Introduction to the capsule network}
\label{5.6}
\subsubsection{What is the capsule network?}
Capsule Network (CapsNet) is introduced by Geoffrey Hinton and his team 
as an advancement over traditional convolutional neural networks (CNNs) \cite{sabour2017dynamic,hinton2018matrix}. 
CapsNet aims to overcome the limitations of CNNs, 
particularly their inability to preserve hierarchical relationships between different features in an image, 
and their inefficacy in dealing with spatial hierarchies in data. 
A capsule is a group of neurons whose activity vector represents the instantiation parameters of a specific type of entity, 
such as an object or an object part. 
Each capsule not only detects the presence of its respective entity but also learns to recognize its various properties 
(e.g., pose, size, texture).
Each capsule vector’s length represents the presence probability of specific entities
in the input image, and its direction encodes the captured features of entities.

\subsubsection{Structure of capsule network}
Capsule Network consists of several layers, each designed to contribute uniquely to the network's functionality:
\begin{itemize}
  \item \textbf{Convolutional Layer:} This layer is similar to that in CNNs and is used for initial feature detection.
  \item \textbf{Primary Capsule Layer:} Converts pixel intensities into a small set of initial capsule outputs, which are then used as inputs to higher-level capsules.
  \item \textbf{Digit Capsule Layer:} This layer consists of higher-level capsules, each representing a more complex entity. It receives inputs from all the capsules in the layer below and uses dynamic routing to decide which lower-level capsules to couple with.
\end{itemize}

\subsubsection{Dynamic routing mechanism}

Dynamic routing is designed to update the connection strengths between capsules across layers based on their agreement. 
The core idea is to allow a lower-level capsule to choose which higher-level capsules to send its output. 
This decision is not hard-coded but learned by the network, which makes capsule networks capable of handling various spatial relationships.
Unlike traditional routing mechanisms in neural networks, 
dynamic routing between capsules allows the network to learn which features are likely to be part of the same entity, 
thus encoding spatial and part-whole hierarchies of the objects. 

The dynamic routing algorithm iteratively adjusts the coupling coefficients between capsules, 
which represent the probability that a lower-level capsule's output should be sent to a higher-level capsule. 
Here is a step-by-step breakdown of the algorithm:
\begin{enumerate}
  \item \textbf{Initialization:} All the log probabilities $b_{ij}$ are initialized to zero. These log probabilities indicate the initial likelihood that the output of capsule $i$ should be sent to capsule $j$.
  
  \item \textbf{Routing iterations:} For each iteration, the algorithm updates the coupling coefficients $c_{ij}$ and the outputs of higher-level capsules. The updates are performed as follows:
  \begin{align*}
      \mathbf{u}_i &= \text{output of capsule } i, \\
      b_{ij} &\leftarrow b_{ij} + \mathbf{u}_i^\mathsf{T} \mathbf{v}_j, \\
      c_{ij} &\leftarrow \frac{\exp(b_{ij})}{\sum_k \exp(b_{ik})},
  \end{align*}
  where $\mathbf{u}_i$ is the output vector of capsule $i$, $\mathbf{v}_j$ is the vector output by capsule $j$ in the previous iteration, and $c_{ij}$ are the updated coupling coefficients calculated by a softmax function over $b_{ij}$.
  
  \item \textbf{Output calculation:} Each higher-level capsule $j$ then computes its new output vector $\mathbf{v}_j$ as a weighted sum of all candidate vectors $\mathbf{u}_i$, where the weights are the coupling coefficients $c_{ij}$:
  \begin{align*}
      \mathbf{v}_j &= \sum_i c_{ij} \mathbf{u}_i.
  \end{align*}
\end{enumerate}

This procedure allows the network to send more information to higher-level capsules that "agree" — meaning their outputs are similar, 
thereby enhancing the network's ability to recognize complex features composed of simpler features robustly.


\end{document}